\documentclass[runningheads]{llncs}
\usepackage{breqn}
\usepackage[T1]{fontenc}
\usepackage{xcolor}
\usepackage{graphicx}
\begin{document}
\title{A metric to compare the anatomy variation between image time series}
%
%
\author{Alphin J Thottupattu \inst{1} \and
Jayanthi Sivaswamy\inst{1}
}
%
%
\institute{International Institute of Information Technology
\email{alphinj.thottupattu@research.iiit.ac.in}\\
}
\maketitle              
\begin{abstract}
Biological processes like growth, aging, and disease progression are generally studied with follow-up scans taken at different time points, i.e., with image time series (TS) based analysis. Comparison between TS representing a biological process of two individuals/populations is of interest. A metric to quantify the difference between TS is desirable for such a comparison. The two TS represent the evolution of two different subject/population average anatomies through two paths. A method to untangle and quantify the path and inter-subject anatomy(shape) difference between the TS is presented in this paper. The proposed metric is a generalized version of Fréchet distance designed to compare curves. The proposed method is evaluated with simulated and adult and fetal neuro templates. Results show that the metric is able to separate and quantify the path and shape differences between TS.

\end{abstract}
\section{Introduction}
Studying natural processes such as growth, disease progression, and other physiological processes often requires imaging at different time points, thus generating an image time series (TS). The images in such TS typically represent a deforming organ of an individual or population average anatomies derived to represent the general trend of a process. Modeling the TS as a continuously deforming image/\textit{shape} though a temporal \textit{path} \cite{acceleration2,GEOREGRESSION2}  helps to directly analyze the deformation happening in the anatomy with time \cite{diff_growth}. However, it is a fact that anatomy and the biological process vary across individuals. When two TS are of individuals from the same population, it is presumed that there is anatomical  (or \textit{shape}) similarity, and for a population-level analysis with a group of TS, the focus is on understanding the \textit{path} difference directly or by mapping to a common space \cite{pennec}. For a pair of TS, both \textit{shape} and \textit{path} difference will significantly contribute towards the difference between them, and it has to be captured by a metric that defines the distance between them. It is more interesting to study these separately when comparing two population average TS. Let us consider comparing normal brain aging trends of two populations. Aging leads to brain anatomy deformations, which can differ across populations. Comparing the aging of two populations requires accounting for the basic shape differences between the two. This paper proposes a novel approach that quantifies shape and path variation separately to define the distance between two-time series, reflecting the time-independent and time-dependent variations. This method can be useful in comparing and understanding brain aging differences across different populations.\\
\\
Image similarity metrics such as SSIM and MSE are popular for image-level comparisons. They are inappropriate for image TS because they fail to quantify spatial and temporal differences separately. The terminology 'path difference' is generally used to compare two 1D curves, where the initial points in both curves are assumed to be the same. Hausdorff distance \cite{HD} is a common metric to measure the distance between two curves, but it does not consider the course of the curves. Dynamic time Warping \cite{dtw} aids in comparing two trajectories of different speeds, but it is also a discrete measure. Fréchet distance \cite{frechet_org} is the standard for comparing two continuous curves or paths. The idea of Fréchet distance is defined in \cite{frechet_org} as the minimum leash length when a person walks with a dog on a leash forward, from start to end. In image TS representing biological processes, the spatial context is as important as the temporal variation, but Fréchet distance (FD) \cite{frechet_poly} is not designed to handle these directly. In general, none of the existing methods for curve comparison are directly applicable to compare and quantify 3D image TS differences.\\
3D-image-based qualitative and quantitative comparison of biological process in the existing literature is limited to characteristics like cerebral volume and dimensions of the brain, which are derived from each TS \cite{Brain_dev,INTERESTING1,population,brain_dev3}; this translates to image level comparison for 3D image TS. Such analysis, however, will not help to separately compare the \textit{shape} variation and \textit{path} variations. Measures like changes in volume and structure/organ dimensions have been reported but cannot capture the non-rigid anatomical changes. For instance, growth trajectories have been compared in \cite{pennec} to study the difference between the human brain in healthy/normal individuals and those with Alzheimer's disease by first mapping the trajectories of two groups of individual follow-up scans to a common space in  \cite{pennec} and then performing a volume change analysis. Such methods are useful for case-specific group analysis, but a metric that defines the distance between two TS will facilitate a more general analysis framework.\\
\subsection{Our Contribution}
The main contribution of this paper is a metric for comparing a pair of TS, which considers both \textit{shape} and \textit{path} variations. To our knowledge, this has not been addressed in the context of group analysis. 
We propose a metric to quantify \textit {path} variation inspired by the idea of FD for curves because FD considers the course of the path, unlike other measures, and defines a single metric to quantify the \textit{path} variation. We also propose a  metric to quantify \textit {shape} variation based on the deformation-based distance between the \textit{shape}s of the individual TS. Proposed \textit{shape} and \textit{path} distance metrics are defined for every point in 3D space. The sum of the average \textit{shape} and \textit{path} distances quantifies the difference between the two TS.
\section{Method}
TS data corresponds to either an individual anatomy variation or an average population anatomy variation with time.
It should be pointed out that affine invariant, intensity normalized image TS are considered in this paper as these factors are separately quantifiable.
We also assume that the two TS to be compared have been acquired from two different subjects (and denoted as $TS_1$ and $TS_2$) for approximately the same time range, with aligned time indices. For example, $TS_1$ and $TS_2$ correspond to two subjects' scans in an age range of ~20-30 years, and one subject is scanned only in even years while the other is in odd years. Hence, lack of temporal correspondence and time range mismatch can make comparing two TS difficult, and the discrete nature of the TS causes the same. We propose to overcome this by deriving a continuous model for each TS as a first step. These models of TS are then compared in a common time interval, and \textit{shape} and \textit{path} distances are defined.
\subsection{Deriving continuous representation of TS}
In computational anatomy, a natural process in the human body is typically modeled as a deformation of an underlying anatomy \cite{diff_growth}. Hence, when a sequence of images of a single individual is represented as a TS, it is modeled either  as an anatomy deforming through a path \cite{GEOREGRESSION2,ours} or with a Kernel-based regression \cite{kernel2}. The latter does not model the path as a function of time; instead, any image point is a function of time. Since we aim to separately quantify the difference in TS in terms of \textit{shape} and \textit{path} variations, we choose a path-based modeling approach. When two TS are compared \textit{shape} and \textit{path}, distance is expected to separately quantify the time-dependant and independent variations, respectively. The \textit{shape} in each TS continuous model is the anatomy/image that can be mapped to any time point in the TS through a \textit{path} defined on that. The \textit{shape} in each TS should represent the same time point, then the distance between them represents time-independent variation between TS. 
\\
To illustrate the proposed method, recent path-based modeling from \cite{ours} has opted to derive the continuous model. To derive a continuous model of a TS in time interval $[t_0,t_n]$, any point in the TS can be selected as \textit{shape}, and the \textit{path} is defined for the selected \textit{shape}. A point somewhere in the middle of $[t_0,t_n]$ is chosen for convenience as the temporal range mismatch can be compensated easily with such modeling. The paths are modeled as diffeomorphic deformations $\phi=\exp(v \cdot \gamma(t))$ where $v$ is a vector field that defines the direction in which each spatial position has to move/deform, and the $\gamma(t)$ controls the rate of change of the deformation with time. As the \textit{shape}($S$) lies towards the middle, say at $t=m$, two paths are defined in $[t_0,m]$ and $[m,t_n]$ to cover the whole time range. The TS is modeled as $\left \{ \left \{ \phi_1(t) \circ  S \right \}_{t_0}^{m},\left \{ \phi_2(t) \circ  S \right \}_{m}^{t_n} \right \}$. The deformations $\phi_1(1)$ and $\phi_2(t)$ correspond to the  paths defined in time intervals $[t_0,m]$ and $[m,t_n]$ respectively.  
\\
 Let us consider two TS, $TS_1$ and $TS_2$ given in time interval $[t_i,t_m]$ and $[t_j,t_n]$ respectively. Let the selected \textit{shape}s be $S_I$ and $S_J$, occurring at $m_I$ and $m_J$. Similarly, $\phi^I_*$ and $\phi^J_*$ correspond to the \textit{ path}s of the two models derived on $S_I$ and $S_J$ respectively, where $_*$ corresponds to the path index($1$ or $2$) throughout the paper. Hence, the continuous representation of $TS_1$ and $TS_2$  are $I(t)$ and $J(t)$ as given in Equation \ref{Tta}-\ref{Ttb}. 
\begin{equation}
I(t)=\begin{cases}
 \phi_1^I(t) \circ S_I \mbox{ for } t\geq m_I, \\ 
  \phi_2^I(t) \circ S_I \mbox{ for } t\leq m_I. 
\end{cases}
\label{Tta}
\end{equation}
\begin{equation}
J(t)=\begin{cases}
 \phi_1(t)^J \circ S_J \mbox{ for } t\geq m_J, \\ 
 \phi_2(t)^J \circ S_J \mbox{ for } t\leq m_J. 
\end{cases}
\label{Ttb}
\end{equation}

\subsection{Temporal alignment of the continuous representations}
\label{time_match}

From Equation \ref{Tta}-\ref{Ttb}, it can be noted that shapes $S_I$ and $S_J$ correspond to time points $m_I$ and $m_J$, respectively. Before any comparative assessment of TS, the models need to be temporally aligned by moving $S_I$ and $S_J$ to the same time point. The \textit{shapes} obtained after alignment are denoted as $\tilde{S}_I$ and $\tilde{S}_J$ defined at $(m_I+m_J)/2$. If $m_I<(m_I+m_J)/2$ and $m_J>(m_I+m_J)/2$ then $\tilde{S}_I$ and $\tilde{S}_J$ are given by Equation \ref{ts1} and \ref{ts2} respectively.
\begin{equation}
\tilde{S}_I=S_I \circ \phi_1^I \left ( \frac{m_J-m_I}{2} \right )
    \label{ts1}
\end{equation}

\begin{equation}
\tilde{S}_J=S_J\circ \phi_2^J \left ( \frac{m_I-m_J}{2} \right )
\label{ts2}
\end{equation}

\begin{figure}[h!]
    \centering
    \includegraphics[width=1\columnwidth]{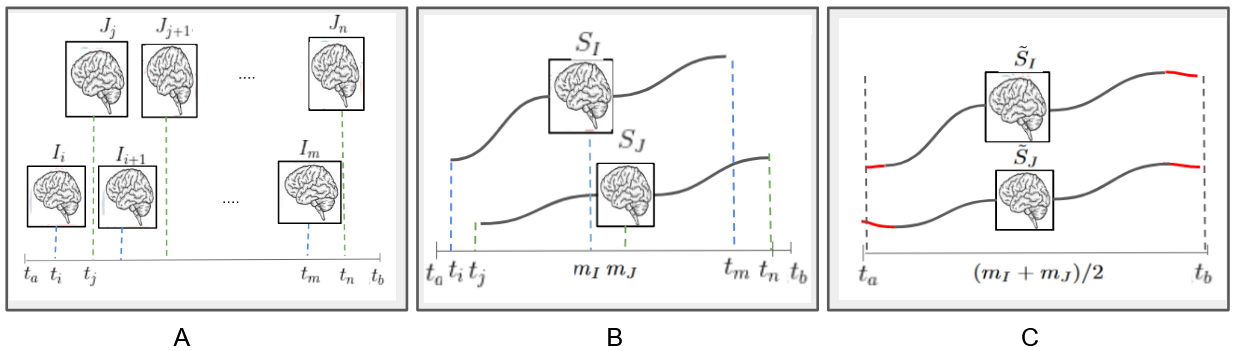}
    \caption{A) Temporally aligned $TS_1,TS_2$, B) Continuous image paths $I(t),J(t)$ , and C) Temporally aligned continuous models with extrapolated deformation(red curve) }
    \label{j}
\end{figure}

 The \textit{shape} distance computation is formulated such that the \textit{shape}s being compared to be at the same time point. To keep \textit{shape}s at $(m_I+m_J)/2$ as shown in Figure \ref{j}.B, the deformations also must be reformulated about the time point $(m_I+m_J)/2$. In Equation \ref{ts1}, $\tilde{S}_I$ is already a deformed version of $S_I$ with a small deformation $\phi_1^I \left ( \frac{m_J-m_I}{2}\right )$, and the same deformation has to be removed from the $\phi_1$ path of the model with $\tilde{S}_I$ as the \textit{shape}. Similarly, other deformations have to be updated, and the updated deformations are given as follows.

\begin{alignat}{1}
     \tilde{\phi}^{I}_1(t)=\phi_1^I(t) \circ -\phi_1^I \left ( \frac{m_J-m_I}{2}  \right )
    \label{c1}
\end{alignat}
\begin{alignat}{1}
        \tilde{\phi}^I_2(t)=\phi_2^I(t) \circ -\phi_1^I \left ( \frac{m_J-m_I}{2}  \right )
    \label{c2}
\end{alignat}

\begin{alignat}{1}
     \tilde{\phi}^{J}_1(t)=\phi_1^J(t) \circ -\phi_1^J \left ( \frac{m_I-m_J}{2}  \right )
    \label{c3}
\end{alignat}
\begin{alignat}{1}
        \tilde{\phi}^J_2(t)=\phi_2^J(t) \circ -\phi_1^J \left ( \frac{m_I-m_J}{2}  \right )
    \label{c4}
\end{alignat}

The continuous models are extrapolated /truncated to the same time interval before comparing the two. Figure \ref{j} shows the steps to be followed to derive the continuous temporally aligned, range-compensated continuous models from the TS, which are used to compute the distance between the pair of TS.

\subsection{Computing the \textit{shape} and \textit{path} distance}
\subsubsection{\textit{Shape} distance:}
 The $\tilde{S}_I$ and $\tilde{S}_J$ in each TS continuous model represents the \textit{shape} corresponding to $TS_1$ and $TS_2$, respectively at the same time point. Hence, the deformation between the $\tilde{S}_I$ and $\tilde{S}_J$ captures the \textit{shape} variation $\phi_S$ between the two TS.  deformation is modeled as  $\phi_S=\exp( \mathbf{V}_S)$, where $\mathbf{V}_S$ represents a stationary velocity field. Then the norm of the vector field $\mathbf{V}_S$ can be directly used to quantify the deformation as given in \cite{logdemon}. The \textit{shape} distance ($d_s$) between $TS_1$ and $TS_2$ is hence defined as \\
 \begin{equation}
     d_s=\left \| {\mathbf{V}_S} \right \|
     \label{shape}
 \end{equation}
 \subsubsection{\textit{Path} distance:} Our aim is to enable the comparison of a pair of TS on a common interval $[t_a,t_b]$, which can be flexibly selected. Extrapolation or truncation may be required, depending on the selected time interval.
 \begin{figure}[ht!]
    \centering
    \includegraphics[scale=0.3]{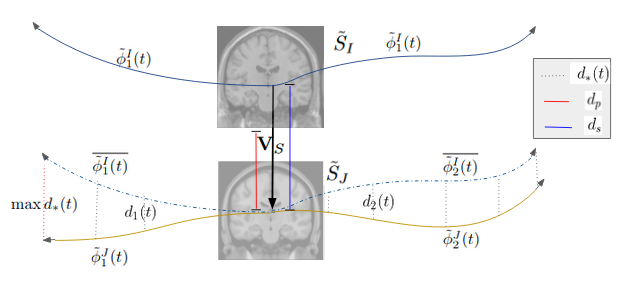}
    \caption{\textit{Shape}($d_s$) and \textit{path}($d_p$) distance computation}  
    \label{proposed}
\end{figure}
 
The \textit{path} distance is defined as the maximum distance between the two paths over the chosen interval. 
A schematic for computing the \textit{path} distance is shown in Figure \ref{proposed}.  The distance between the paths 
has to be computed after separating the \textit{shape} distance between the paths. When the two TS are modeled with the same \textit{shape}, then the only distance between the TS will be the \textit{path} distance ($d_p$). Hence, we map the paths to either $\tilde{S}_I$ or $\tilde{S}_J$ via parallel transport to force $d_s=0$.  Here we consider $\tilde{S}_J$ as the reference to define the \textit{shape} and \textit{path} distance. Hence,  the paths ${\tilde{\phi}^{I}_1(t)}$ and ${\tilde{\phi}^{I}_2(t)}$ are transferred to $\tilde{S}_J$ via parallel transport \cite{Parelle_Transport} through $\mathbf{V}_S$ to get $\overline{\tilde{\phi}^{I}_1(t)}$ and $\overline{\tilde{\phi}^{I}_2(t)}$. The $J(t)$ paths $\tilde{\phi}^{J}_1(t)$ and $\tilde{\phi}^{J}_2(t)$ and the transferred  paths are defined on $\tilde{S}_J$. 
\\
Let the path difference be denoted as $\left \{ d_1(t),d_2(t) \right \}$ where $d_1(t)$ corresponds to the distance in $[t_a,m]$ and $d_2(t)$ corresponds to the distance defined in $(m,t_b]$. Since the path is modeled with vector fields $v \cdot \gamma(t)$, we use the norm of the difference between the vector fields in $I(t)$ and $J(t)$ models to compute the \textit{path} distance $d_*(t)$ as follows. 
\begin{equation}
d_*(t)=\left \|v^I_* \cdot \gamma_I(t)-v^J_* \cdot \gamma_J(t) \right \|
\end{equation}
Finally, the net difference between the paths has to be derived. 
We follow the Fréchet distance formulation for this purpose. By definition, Fréchet distance is the shortest leash with which a man and his dog on the leash can complete a journey. Fréchet distance helps to quantify the similarity by considering the course of the paths. Inspired by the Fréchet distance formulation, the maximum distance $\max{ d_*(t)}$ in $t=[t_a,t_b]$ at each spatial position is computed first. The distance, $\max{ d_*(t)}$ is not defined on $\tilde{S}_J$, hence it is transported to  $\tilde{S}_J$ via $\overline{\tilde{\phi}^{I}_*(t)}$ to get $\overline{\max \left \{ d_{*}(t) \right \}_{t_a}^{t_b} }$. In Figure \ref{proposed} $d_p$ corresponds to \textit{path} distance, and it is given as
\begin{equation}
    d_p=\overline{\max \left \{ d_{*}(t) \right \}_{t_a}^{t_b} }
\end{equation}
\\
 The total distance between the two TS ($D$) is defined in Equation \ref{metric} as a combination of \textit{shape} and \textit{path} variation. The sum of $d_s$ and $d_p$ gives the distance between the TS. Both  \textit{shape} and \textit{path} distances satisfy the distance properties; hence, $D$ also defines a distance that satisfies all distance properties.
\begin{equation}
D=d_s+d_p=\left \| {\mathbf{V}_S} \right \|+\left \| \overline{\max \left \{ d_{*}(t) \right \}_{t_a}^{t_b} }\right \|    
\label{metric}
\end{equation}

\section{Results}
A variety of experiments were done to validate the proposed method. We believe the proposed method is the first attempt towards separating the \textit{shape} and \textit{path} distance between two TS. Hence, bench-marking was not possible.

\subsection{Implementation Details}
If the TS under consideration is longitudinal data, then \textit{shape} $S$ can correspond to any point in the TS, as the same subject is scanned at different time points. If the TS ithe s population average image (i.,e. a template) at each time point, then $S$ is found by averaging all samples in the TS. Now $S$ represents the anatomy that normalizes all inter-subject and temporal variation. In both cases, an $S$ is not preferred to lie at the end of the time range. This constraint in modeling helps to perform time range matching of the two TS. It also demands a two-piece path modeling which is helpful in handling a complex path as a diffeomorphic deformation. 
\subsection{Simulated data-based experiment}
\begin{figure}[ht!]
    \centering
    \includegraphics[scale=0.38]{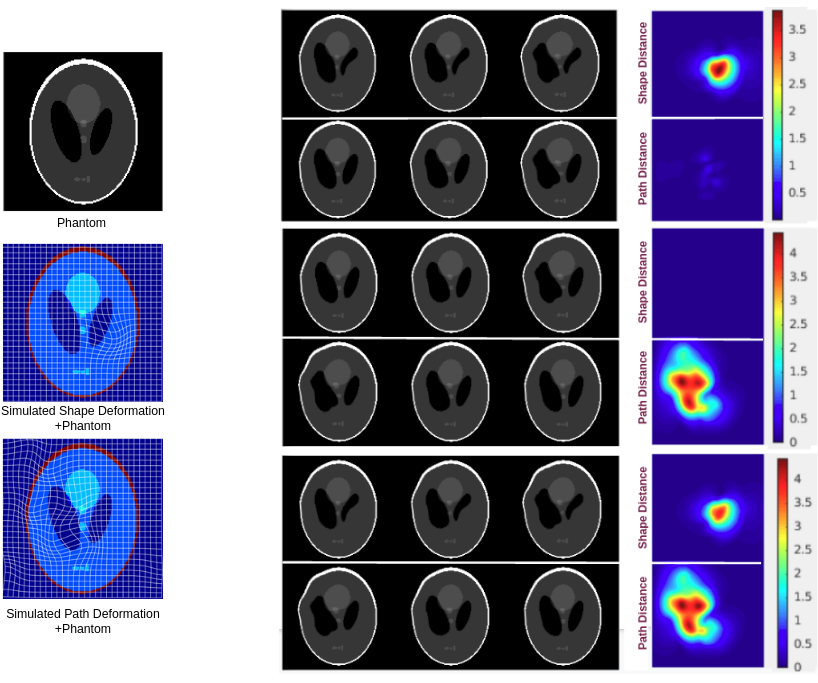}
    \caption{Shepp Logan-based validation of the proposed method to quantify the path distance. Column 1: phantom and the deformations fields for shape and path Column 2: three sets of TS used in the validation; column 4: \textit{Shape} and \textit{path} distance maps}
    \label{sim}
\end{figure}

In order to understand how well the proposed method separates and quantifies time-dependent and independent distances, a simulation experiment was done with three sample TS pairs which were constructed by deforming a Shepp-Logan phantom with simulated path and shape deformation as shown in the first column of Figure \ref{sim}. The first set of TS (rows 1-2) was constructed such that they differed only by shape, while the second set of TS (rows 3-4) was constructed to differ only by the path, and finally, the last set of TS ( rows 5-6) was constructed to differ in terms of both shape and path. 
To generate the second set (rows 3-4), two mutually inverse paths were constructed using the path deformation on the phantom image. The third set (rows 5-6) was constructed with different source images; one was the original phantom image, and the other was the shape-deformed phantom image. Inverse paths were applied to these images to construct the TS pair. 
The last column in Figure \ref{sim} displays a heat map for each set of the computed shape($d_s$) and path($d_p$) distance values. For the first set, the \textit{path} variation is negligible, and \textit{shape} variation is maximum and vice versa for the second set. For the last pair of TS, both \textit{shape} and \textit{path} variations are observed. This observation is in line with the expected results. Hence, this experiment validates the proposed method's ability to separate the time-dependent and independent distances between a pair of TS.
\subsection{Aging data-based experiment}
\begin{figure}[ht!]
    \centering
    \includegraphics[scale=0.3]{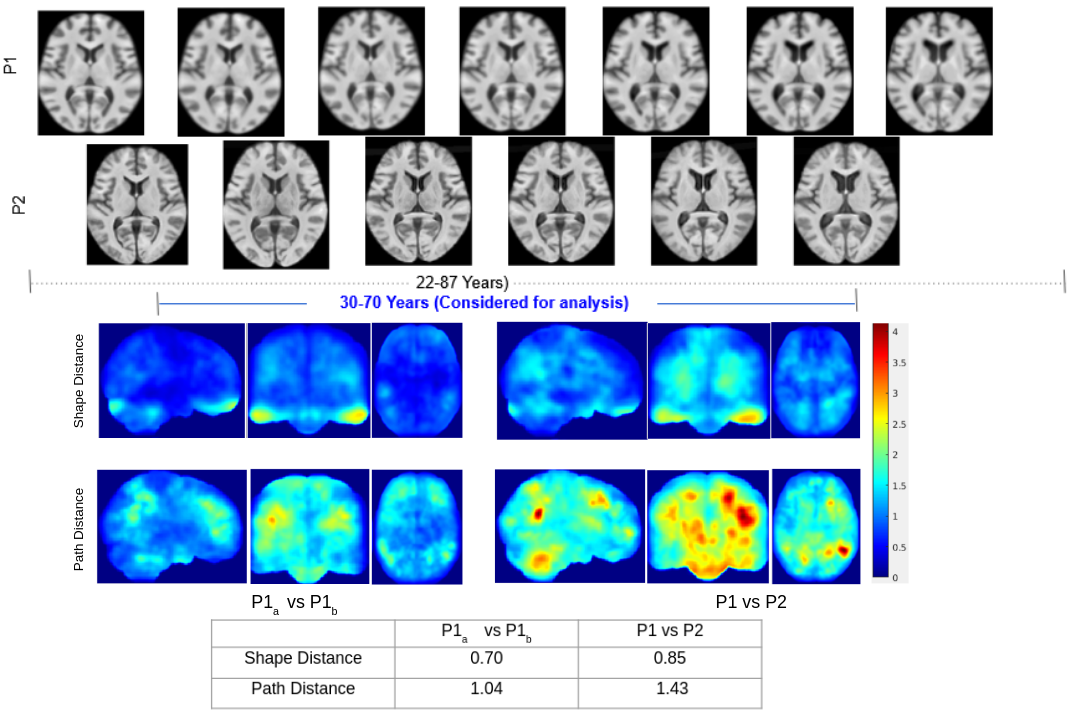}
    \caption{Validation results on TS of templates of adults from two different populations. Sample time points of TS considered for the experiment are shown in 1-2 rows. Row 3-4 shows the 3D visualization of the distance between two TS for intra-population ($P1_a$ vs. $P1_b$) and inter-population ($P1$ vs. $P2$). Last row: 3D average \textit{shape} and \textit{path} distances}
    \label{a}
\end{figure}

The second experiment is with real data, specifically 3D brain templates at different age points, which arise in brain aging studies. 
The aging process and brain anatomy are expected to vary across two different populations \cite{population}. Hence, inter-population distances are expected to be larger than intra-population distances; these hypotheses are evaluated with the proposed metric. Datasets drawn from the Caucasian (Neurodev \cite{neurodev}) and Japanese (AOBA \cite{aoba} populations were used for the inter-population study in this experiment. The age range of subjects was 22-87 years for the former and 25-75 years for the latter. The intra-population TS pair was constructed from the Neurodev data by sampling the data at odd ($P1_a$) and even ($P1_b$) time indices. This was possible because Neurodev templates were defined every 5 years, whereas AOBA templates are only available for every decade. Hence, intra-population analysis was not done with AOBA. \textit{Shape} and \textit{path} distance were computed for inter and intra-population TS pairs and are presented in a Table in Figure \ref{a}. Notably, both shape and path distances are higher for inter-population (column 2) than intra-population TS pairs. The time interval considered for analysis was 30-70 years, as the time ranges differ for the two TS. Sample time points of each TS considered in this analysis are shown in Figure \ref{a}.   
The 3D visualization of \textit{shape} and \textit{path} distances are shown in Figure \ref{a} in three canonical planes for a better understanding of the spatial distribution of the distance. It can be observed that both the shape and path 
distances are smaller within a population relative to across populations; this inter-population difference appears to be primarily due to the \textit{path} difference rather than the \textit{shape} difference, which is consistent with the results listed in the Table.
 
\begin{figure}[ht!]
    \centering
    \includegraphics[scale=0.28]{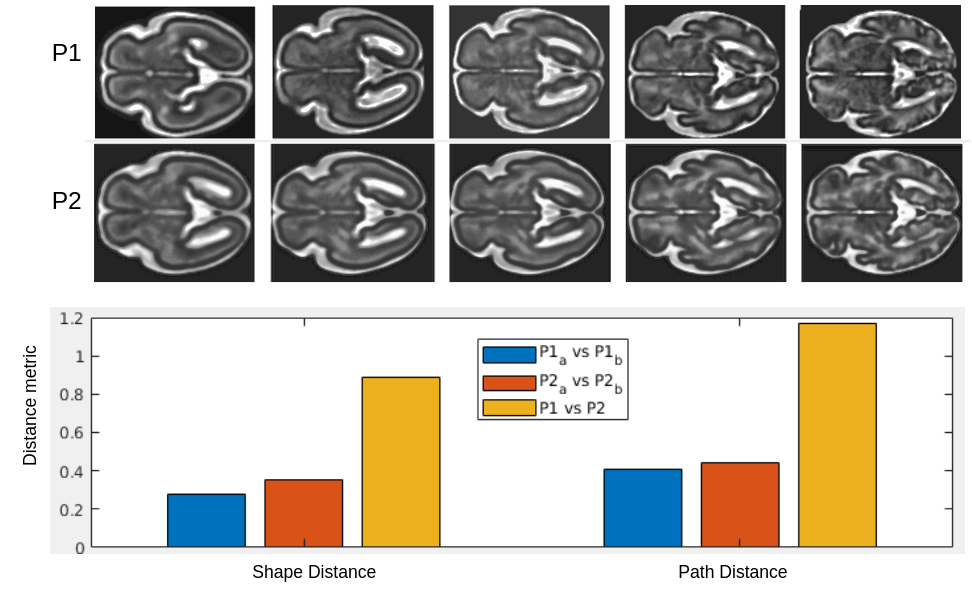}
    \caption{ Validation results of templates of the fetal brain from two populations. The samples of population average templates for the Caucasian and Chinese populations are shown in rows 1-2. The average distance plots for the intra- ($P1_a$ vs. $P1_b$) and inter-population ($P1$ vs. $P2$) study are shown at the bottom.}
    \label{fb1}
\end{figure}

\begin{figure}[ht!]
    \centering
    \includegraphics[scale=0.7]{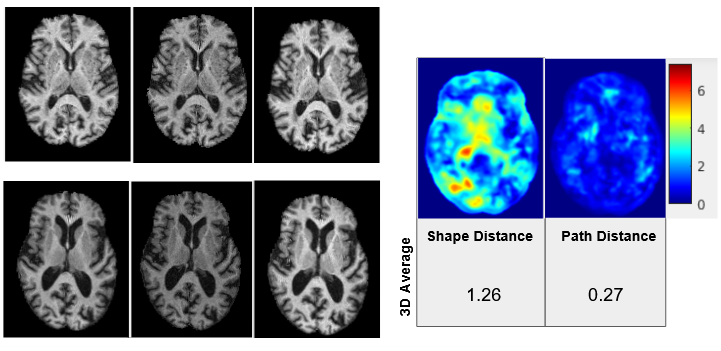}
    \caption{Validation results on a longitudinal pair of TS. Right: First and second row shows 2D slices of TS from two different subjects(78-83 years and 78-82 years). Left: \textit{Shape} and \textit{path} distance maps of the corresponding slice along with 3D averaged values}
    \label{fb2}
\end{figure}

 A third experiment was done with fetal brain datasets as the developmental changes are large, unlike in the case of the adult brain considered in the previous experiment. Caucasian and Chinese populations were considered for this purpose. Scans of subjects aged 23-35 weeks were used for both populations. The fetal templates for the Caucasian population were from CRL database \cite{cau} while those of the Chinese population were from FBA \cite{chinese} database. As both the TS were well sampled, the intra-population TS were generated in the same manner as in the previous experiment. A few sample points of the two TS are shown in Figure \ref{fb1}. $P1_a$ and $P1_b$ constructed from Causian($P1$) and $P2_a$ vs $P2_b$ constructed from Chinese($P2$) populations. Two intra-population cases ($P1_a$ vs $P1_b$, and $P2_a$ vs $P2_b$) and one inter-population ($P1$ vs $P2$) case were compared.\\
 The \textit{shape} and \textit{path} distances are plotted separately for each pair of TS in Figure \ref{fb1}. It is notable that the  \textit{shape} and \textit{path} distances are almost the same for the intra-population pair, whereas the variation is much higher for inter-population pairs. Once again, \textit{Path} distance is higher and is the major contributor to the total distance between two population TS.
Finally, the proposed method was also evaluated on a pair of longitudinal TS acquired from two normal individuals at fairly short intervals of 78-83 years and 78-82 years from the same population \cite{oasis}. The cross-sectional \textit{shape} variation shown in colour maps in Figure \ref{fb2} appears to be more than the \textit{path} variation for the two subjects. This is the opposite of the result for TS covering a wider age range. This trend is logical as the aging effect over a short time span is likely to be much less across subjects from the same population than morphological variation.  
\section{Discussion and Conclusion}
A metric that enables disentangling of the \textit{shape} and \textit{path} variation and helps quantify the difference between a pair of TS is proposed in this paper. The proposed metric is an affine invariant and time interval mismatch-compensated metric. The idea of \textit{shape} variation in the proposed metric is more relevant when the TS under consideration are from cohorts from different populations. As the course of the path is considered in the quantification of the \textit{path} variation, the intra-population TS path variation can be analyzed. For example, one can study the \textit{path} difference in the growth pattern among the elderly (50-80 years) versus the young (20-50) within a population. This was done in our second experiment, and the distance for Caucasians was found to be 1.8, while it is 1.4 for Japanese. This suggests that the temporal variations are faster in Caucasians than in Japanese after adulthood. Whereas the difference across populations is much lower for the fetal brain. Such analysis opens up the opportunity to better understand the reason behind such trends from young to elderly and across populations. 
The main goal of longitudinal data-based group analysis is to understand the general trend. Our work enables approaching the problem via a joint statistical analysis of 4D data (TS of 3D images). A metric to quantify the distance between a pair of TS can also help derive an average TS model from a set of TS as done for 1D-3D objects.
There are some limitations with regard to the proposed metric. It cannot handle large temporal mismatches. Further, the metric accuracy is totally dependent on the accuracy of the computed deformations. This is relevant to inter-subject TS analysis registration in this scenario is generally error-prone that too for a complex structure such as the brain. 

\end{document}